\pdfoutput=1

\documentclass[11pt]{article}

\usepackage[]{acl}
\usepackage{times}
\usepackage{latexsym}

\usepackage[T1]{fontenc}

\usepackage[utf8]{inputenc}

\usepackage{microtype}
\usepackage{times,latexsym}
\usepackage{url}
\usepackage{times}
\usepackage{latexsym}
\usepackage{array}
\usepackage{tikz}
\usepackage{soul}
\usepackage{caption, floatrow}

\usepackage{times}
\usepackage{latexsym}
\usepackage{graphicx}
\usepackage{multirow}
\usepackage{amssymb}  
\usepackage{amsthm}
\usepackage{amsmath}
\usepackage[official]{eurosym}
\newcommand{\Tref}[1]{Table~\ref{#1}}
\newcommand{\Eref}[1]{Eq.~(\ref{#1})}
\newcommand{\Fref}[1]{Figure~\ref{#1}}

\newcommand{\Sref}[1]{Section~\ref{#1}}
\usepackage{arydshln}
\usepackage{algorithm}
\usepackage{algorithmic}
\usepackage{graphicx}
\usepackage{booktabs}
\usepackage{calrsfs}
\usepackage{array}
\usepackage[T1]{fontenc}
\usepackage{subcaption,booktabs}
\usepackage{CJKutf8}
\usepackage[OT2, T1]{fontenc}
\usepackage[russian, english]{babel}
\usepackage[T1]{fontenc}
\usepackage{adjustbox}
\usepackage{fdsymbol}
\usepackage{paralist}
 
\usepackage{arydshln}
\usepackage[algo2e,linesnumbered,ruled,vlined]{algorithm2e}

\newcommand*\samethanks[1][\value{footnote}]{\footnotemark[#1]}

\title{Evidentiality-guided Generation for Knowledge-Intensive NLP Tasks}

\author{ Akari Asai\thanks{~~Work done in part while at the Allen Institute for AI}  \\
University of Washington \\
\And
  Matt Gardner\samethanks \\
  Microsoft Semantic Machines \\
  \\
  \texttt{\{akari,hannaneh\}@cs.washington.edu} \\
  \texttt{mattgardner@microsoft.com}
   \And 
   Hannaneh Hajishirzi \\
  University of Washington \\
  Allen Institute for AI \\
  } 

\begin{document}
\maketitle
\begin{abstract}
Retrieval-augmented generation models have shown state-of-the-art performance across many knowledge-intensive NLP tasks such as open-domain question answering and fact verification.
{These models are trained to generate a final output given retrieved passages that can be irrelevant to an input query, leading to learning spurious cues or memorization. }
This work introduces a method to incorporate  {\it evidentiality} of passages---whether a passage contains correct evidence to support the output---into training the generator.  
We introduce a multi-task learning framework to jointly generate the final output and predict the {\it evidentiality} of each passage. 
{Furthermore, we introduce a new task-agnostic method for obtaining high-quality {\it silver} evidentiality labels, addressing the issues of gold evidentiality labels being unavailable in most domains.}
Our experiments on five datasets across three knowledge-intensive tasks show that our new evidentiality-guided generator significantly outperforms its direct counterpart on all of them, and advances the state of the art on three of them. 
{Our analysis shows that the  multi-task learning and silver evidentiality mining play key roles.}\footnote{Our code is available at \url{https://github.com/AkariAsai/evidentiality_qa}.}

\end{abstract}

\section{Introduction}
\begin{figure}[t!]
  \includegraphics[width=0.95\linewidth]{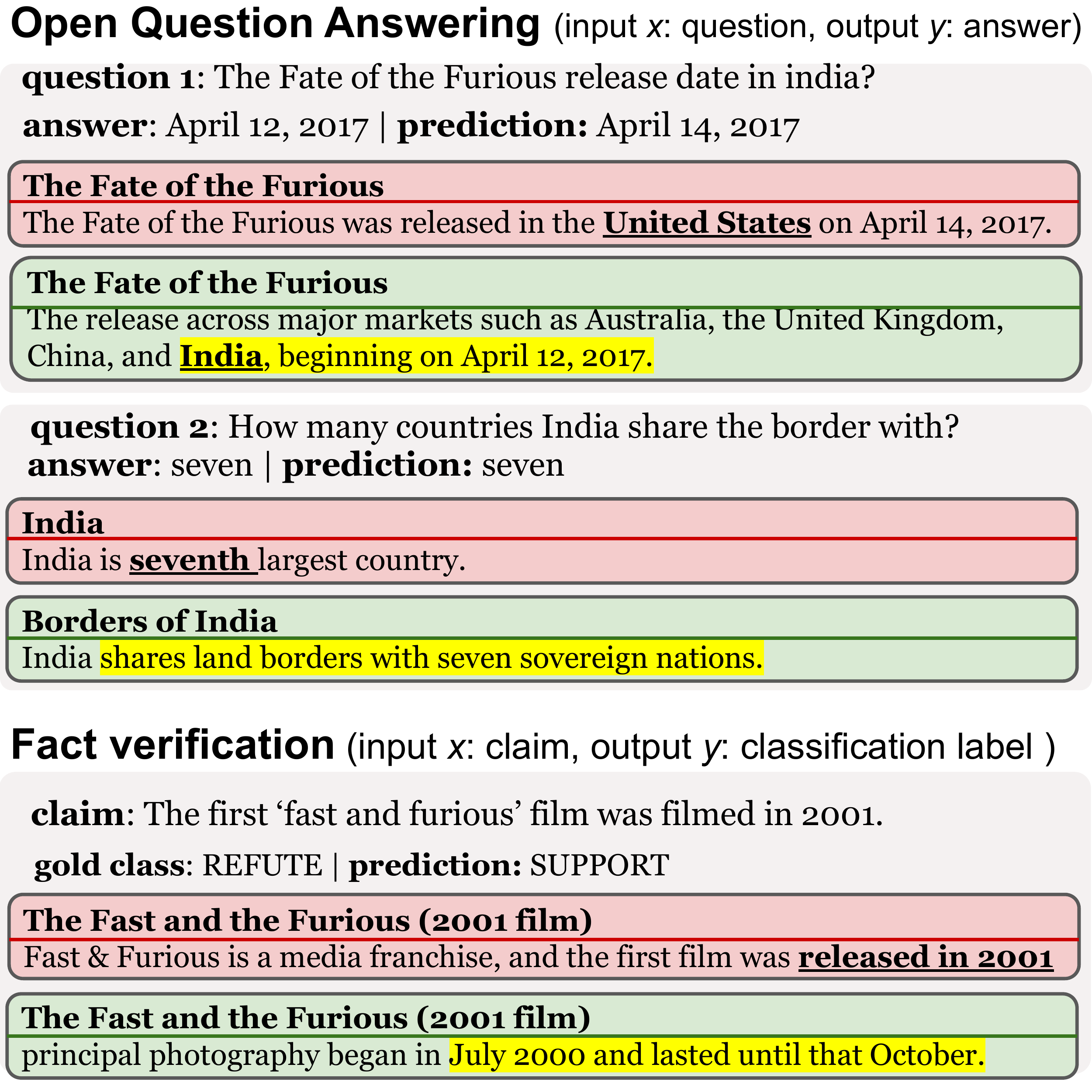}
  \caption{Examples where a trained generator ignores the evidential passages (evidentiality-{\it positive} passages; green rounded rectangles) and makes incorrect predictions from passages that do not provide sufficient evidence (evidentiality-{\it negative} passages; red rounded rectangles). The highlighted part indicates the supporting evidence.}
  \label{img:teaser}
\end{figure}
Knowledge-intensive tasks, including open-domain Question Answering (QA) and fact verification, {require evidence passages related to an input query to be retrieved from a large collection of passages (e.g., Wikipedia). }
Recently, most successful methods use retrieval-augmented generation~\cite{lewis2020retrieval,izacard2020leveraging}, which is a pipeline approach of first training a retriever model~\cite{karpukhin2020dense} for retrieving passages and then independently training a generator model~\cite{lewis2019bart,JMLR:v21:20-074} given the passages. 

Ideally, a model should generate a correct answer given the information presented in {\it evidential} passages~\cite{lee-etal-2021-robustifying} that correctly support the answer and should not be distracted by other passages, even when they happen to contain a string close to the gold answer. However, the disjoint training process in the prior work disregards the evidentiality of passages, leading to  generation models that ignore retrieved passages, leverage spurious cues, and generate hallucinations when the context is not evident~\cite{longpre2021entitybased,xu2021attention}. In particular, incorrectly-retrieved passages with high lexical overlap to the query can mislead the answer generator (the first example in \Fref{img:teaser}).
Adopting heuristics such as answer string matching~\cite{chen2017reading} to train a QA model with passages containing the target strings can partially solve this problem for some QA tasks. Still, these passages with answer strings might lack evidence (the second example in \Fref{img:teaser}). 
What is more, such heuristics cannot be applied for open-ended generation or classification tasks (the third example in \Fref{img:teaser}).  

In this paper, we introduce a multi-task training framework of answer generation and evidentiality prediction, which is an auxiliary task to predict if a passage provides evidence relevant to the task (evidentiality-{\it positive} passages; green passages in \Fref{img:teaser}) or not (evidentiality-{\it negative} passage; red passages in \Fref{img:teaser}). 
Since most existing datasets do not provide evidentiality labels, we introduce a new task-agnostic approach for mining {\it silver} evidentiality annotations. 

Specifically, we train an evidentiality labeling model that takes an input query, a gold output and a single passage and predicts if the passage supports the gold output or not.  
{After training, the evidentiality labeling model predicts the {\it silver} evidentiality labels of all of the passages used for the multi-task training. }
To supervise this evidentiality labeling, we use a combination of {partially available} gold passage annotations and data collected by a novel leave-one-out generation approach. This leverages a trained generator model and evaluates the relevance of each passage to a query through the correctness of the generated output when the passage is removed from the pool of retrieved passages. 
{
Unlike prior multi-task learning work in QA relying on available annotated data~\cite{lee-etal-2021-robustifying,nishida2019qfe} or heuristics such as answer string matching to label pseudo evidentiality~\cite{fajcik-etal-2021-r2-d2}, our approach is applicable to diverse downstream tasks, where we cannot use additional annotations or heuristics. 
Our evidentiality mining approach for high-quality silver labels can be applied to diverse NLP tasks, and our auxiliary task has a new purpose of evaluating passage evidentiality suitable for the open-retrieval. 
}

We run experiments across representative knowledge-intensive tasks: open-domain QA (Natural Questions Open; \citealp{kwiatkowski2019natural}, TriviaQA unfiltered;~\citealp{JoshiTriviaQA2017}), fact verification (FaVIQ Ambig;~\citealp{park2021faviq}, FEVER;~\citealp{thorne-etal-2018-fever}) and {knowledge-enhanced} dialogue \cite[Wizard of Wikipedia;][]{dinan2018wizard}. 
Our experiments show large performance improvements across all datasets over the direct counterpart, FiD~\cite{izacard2020leveraging}. Moreover, on the latter two tasks, our model outperforms all previously published models, advancing state of the art on FaVIQ-Ambig, FEVER and Wizard of Wikipedia.
Further human evaluations find that the evidentiality labeling model yields 95\% accuracy, and often correctly identifies negative passages spuriously containing answer strings.
Our analysis shows that both multi-task learning and silver evidentiality mining contribute to the improvement, helping the generator learn to focus on the more relevant passages.
\begin{figure*}[ht!]
  \includegraphics[width=0.95\linewidth]{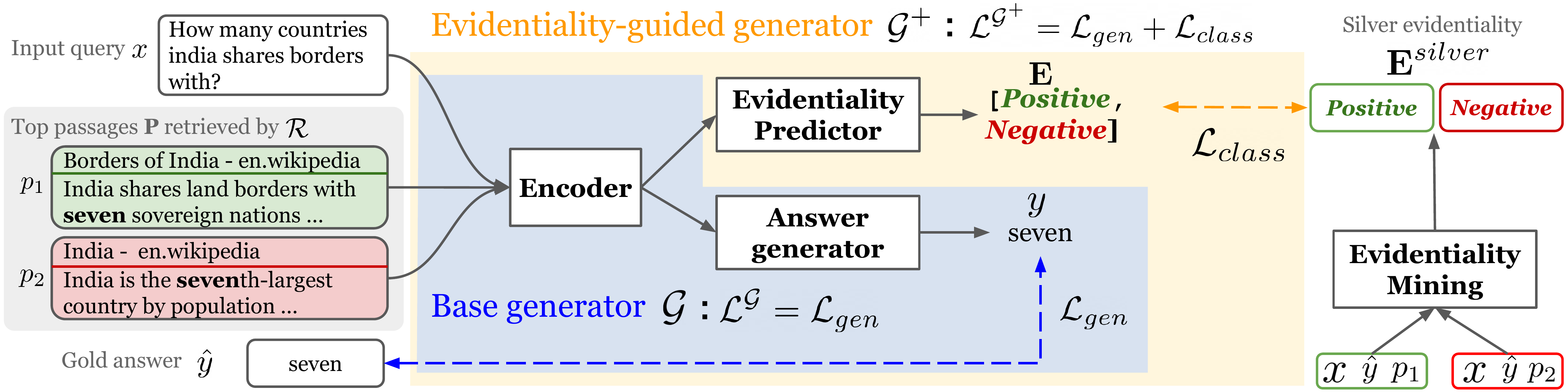}
  \caption{Overview of our proposed framework. The components inside the blue rectangle is a base generator $\mathcal{G}$ and our evidentiality-guided generator is the area inside the yellow rectangle. The straight arrows represent the input-output flow, and the dashed arrows indicate the losses. 
  }
  \label{img:overview}
\end{figure*}

\section{Method}
\subsection{Overview}
\paragraph{Problem.} 
Knowledge-intensive tasks (e.g., open-domain QA, fact checking) are designed to retrieve evidence passages related to an input query $x$ given a large collection of passages such as Wikipedia.
Most successful previous work in this domain uses a retrieval-augmented generation framework such as Fusion-in-Decoder~\cite[FiD;][]{izacard2020leveraging} that consists of two components: a retriever model $\mathcal{R}$ and a generator model $\mathcal{G}$. 
The retriever model $\mathcal{R}$ is trained to retrieve a set of passages $\mathbf{P} = \{p_1, p_2, \ldots, p_i, \ldots, p_N\}$ with the highest top $N$ relevance score for each training query $x$: $\mathbf{P} = \mathcal{R}(x)$. 
The base generator model $\mathcal{G}$ (\Sref{sec:fid}) is then trained to generate the final output $y$ given an input query and the top retrieved passages: $ y = \mathcal{G}(x, \mathbf{P})$. 

Our analysis (Appendix in \Sref{sec:fid_analysis}) shows that a base generator $\mathcal{G}$ trained in this manner often generates the answers from passages ranked high by the retriever, which are not necessarily the correct evidence passages.
Our goal is to build a model that recognizes the {\it evidentiality} of each passage and generates answers based only on passages that contain relevant evidence.
{
We define passages with evidence relevant to the task as {\it positive} and passages without evidence as {\it negative}, even if they happen to include some spurious cues a model can exploit (e.g., a gold answer string for QA). 
}

\paragraph{Method overview.} 
Our method extends the retrieval-augmented generation paradigm by improving the generator $\mathcal{G}$ to generate answers from passages with correct evidence. 
We train our new evidentiality-guided generator $\mathcal{G}^{+}$ using a multi-task learning framework, sketched in \Fref{img:overview}. 
Specifically, given an input query $x$, we combine the generation of the correct answer $\hat{y}$ with the prediction of binary evidentiality labels for each passage in $\mathbf{P}$ used for training: $\hat{\mathbf{E}}= \{\hat{e}_1, \hat{e}_2, \ldots, \hat{e}_i, \ldots, \hat{e}_N\}$.

It is challenging to obtain {\it gold} evidentiality labels $\hat{\mathbf{E}}$ for many tasks. Most datasets are curated with only query-answer annotations $(x, \hat{y})${, or cover subsets of gold passages existing in the large collection of passages, and {considering those original gold passages as only positive passages may result in many false negative passages with correct evidence.} }
Therefore, we heuristically obtain {\it silver} evidentiality data $\mathbf{E}^{silver}$ (\S\ref{sec:data}) by training an evidentiality labeling model $\mathcal{M}$ that assigns a silver evidentiality label $e^{silver}_i$ to each passage $p_i$ given the query $x$ {and the gold output $\hat{y}$}.
In order to find gold evidence passages to train $\mathcal{M}$, we introduce a new approach to evaluate the relevance of passages in generating the correct answer by leaving one passage at a time in answer generation (called {leave-one-out generation}, sketched in \Fref{img:evidentiality}).  
{We mine new gold passages for the target task, and} train $\mathcal{M}$ using the mixture of {partially available gold evidence passage data and newly mined data}. 
{After training, we run $\mathcal{M}$ on all the training data $(x, \mathbf{P}, \hat{y})$ to obtain $\mathbf{E}^{silver}$.}

Finally, we describe {auxiliary multi-task learning} (sketched in \Fref{img:overview}) using $(x, \hat{y})$ and the newly mined silver evidentiality data $\mathbf{E}^{silver}$ in \Sref{sec:loss}.
Our evidentiality-guided generator $\mathcal{G}^{+}$ learns to simultaneously predict the probabilities of output sequences $y$ and evidentiality for {all of the input passages} $\mathbf{E}$.
 
\subsection{Base Generator $\mathcal{G}$}
\label{sec:fid}
We use FiD~\cite{izacard2020leveraging}, a state-of-the-art retrieval-augmented generation model, as our base generator model $\mathcal{G}$. 
We include a high-level summary of the model for clarity, referring the reader to \citet{izacard2020leveraging} for more details.

\paragraph{Encoder.}
We first encode the input query and passages using a pre-trained T5~\cite{JMLR:v21:20-074} encoder. 
The input query $x$ is prepended to each passage, and the encoder encodes each of $N$ passages independently. 
Formally, we transform passage $p_i$ into $\mathbf{p}_i \in \mathbb{R}^{L \times h}$, where $L$ is the input text length and $h$ is a hidden size. 

\paragraph{Answer generator.}
 $\tilde{\mathbf{P}}$ is an input summary representation, formed by concatenating $\mathbf{p}_1, \dots, \mathbf{p}_N$. 
The answer generator takes $\tilde{\mathbf{P}}$ and outputs the final answer autoregressively. 
Specifically, it outputs the sequence probability for $y$ as follows:  
\begin{equation*}
    P(y | x, \tilde{\mathbf{P}} ) = \prod_{j=1}^{T}p(y_j | y_{< j}, x, \tilde{\mathbf{P}}). 
\end{equation*}
where $y_j$ denotes the $j$th token of the generated output $y$ and $T$ is the length of the final output. 
The generator is based on the T5 architecture and uses cross attentions to model the interactions between retrieved passages. 

\subsection{Mining Silver Evidentiality $\mathbf{E}^{silver}$}
\label{sec:data}
{As discussed above, evidentiality labels are unavailable in most of the datasets, and even in some datasets with gold evidence annotations such as Natural Questions~\cite{kwiatkowski2019natural}, it only covers subsets of gold passages from certain articles. }
To overcome these limitations, we introduce an evidentiality labeling model $\mathcal{M}$, which computes the probability that a paragraph $p_i$ contains evidence for an input $x$, given the correct answer $\hat{y}$: $p({e}^{silver}_i | x, p_i, \hat{y})$.  
We use a RoBERTa~\cite{liu2019roberta}-based binary classification model for $\mathcal{M}$.
This model is trained using gold evidentiality annotations when those are partially available, or using labels obtained from a new heuristic mining approach described below. 
Finally, we use the trained evidentiality labeling model to generate {\it silver} evidentiality labels for all of the passages included in the training data.  

\begin{figure}[t!]
  \includegraphics[width=0.95\linewidth]{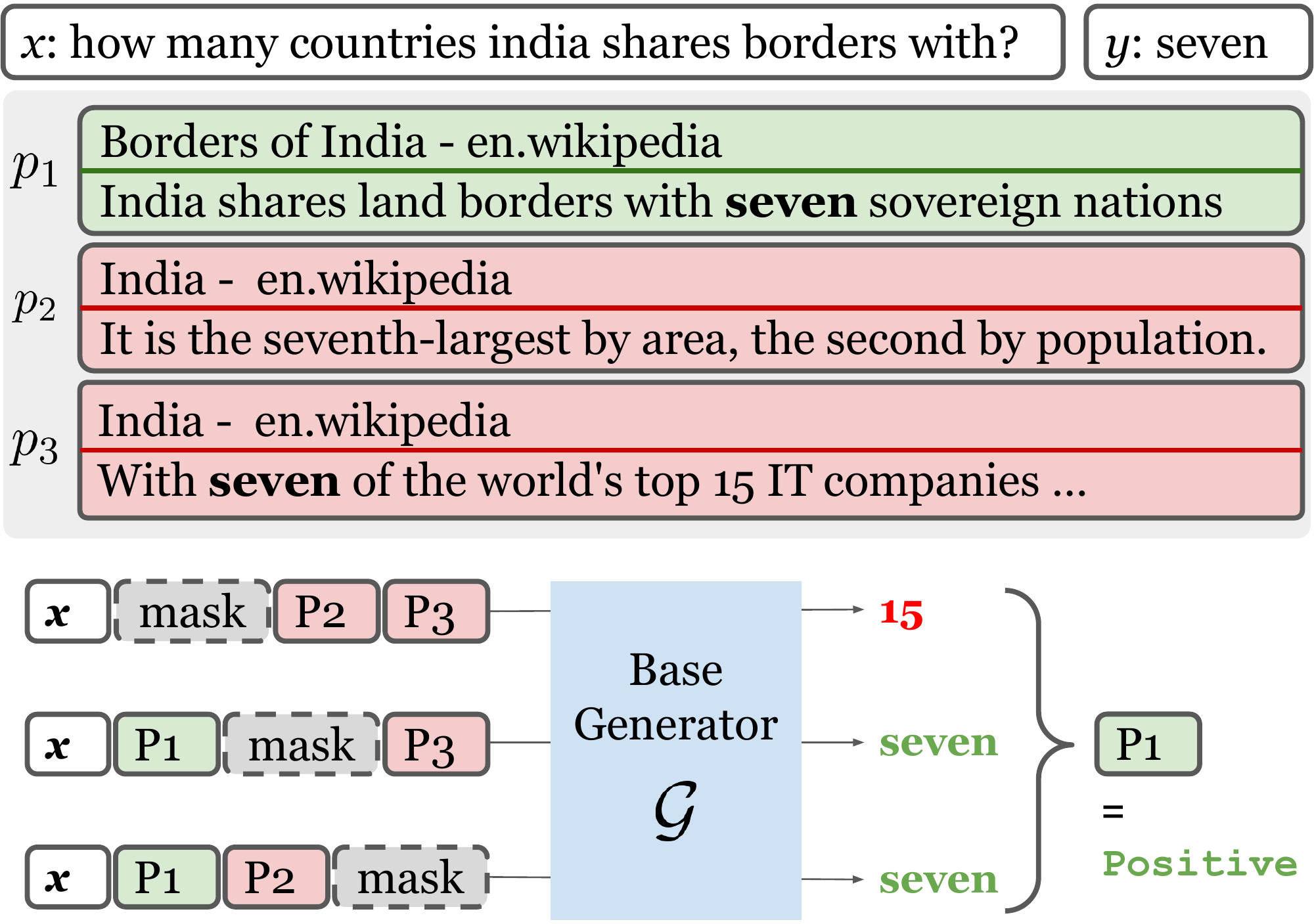}
  \caption{{Overview and examples of our leave-one-out generation to find new positive and negative examples. We mask (remove) one passage at each iteration.}}
  \label{img:evidentiality}
\end{figure}

\paragraph{{Leave-one-out generation.} }
{
To precisely identify gold passages with correct evidence when a target dataset only has input-output annotations, our leave-one-out generation approach (sketched in \Fref{img:evidentiality}) leverages a trained base generator model and uses its predictions to estimate the relevance to the query of the passage. }
Specifically, we feed an input query $x$ and retrieved passages $\mathbf{P}$ to our trained base generator for $N$ times, where we mask the $i$th passage in the $i$th iteration to evaluate if the model can still generate the correct answer without the information presented in $i$th passage.
We consider $i$th passage $positive$ if the model fails to generate $\hat{y}$ when and only when $i$th passage is masked. 
We also consider $i$th passage $negative$ if the model succeeds in generating $\hat{y}$ when and only when $i$th passage is masked---this means that the $i$th passage confuses the model. 
{
This approach may not find all of the gold evidence passages when there are multiple gold passages in $\mathbf{P}$ or the answers are memorized during fine-tuning of the base generator. Yet, we found that we can mine a sufficient number of high-quality gold passages using our approach to quickly adapt the evidentiality labeling model $\mathcal{M}$ to a new task.
}
In our experiments, we combine the gold evidentiality data (i.e., long answers) from Natural Questions with task-specific leave-one-out data to train a separate evidentiality model $\mathcal{M}$ for each task.
{See the details of the data mining for each task in Appendix. }

\subsection{Multi-task Learning with $\mathbf{E}^{silver}$}
Our generator $\mathcal{G}^+$ shares a similar, T5-based encoder-decoder architecture as the base generator, but we have an additional decoder that is used for the evidentiality prediction.
We train $\mathcal{G}^+$ with a multi-task objective given the originally available data $(x, \mathbf{P}, \hat{y})$ and newly mined $\mathbf{E}^{silver}$.

\paragraph{Evidentiality predictor. } 
The evidentiality predictor predicts the evidentiality of each passage. 
{Similarly to} the answer generator, we use the T5 decoder architecture for the classifier. 
Our evidentiality predictor generates the evidentiality $e_i$ given encoded passage representation $\mathbf{p_i}$: $p(e_i | q, \mathbf{p_i} )$. The evidentiality predictor in $\mathcal{G}^+$ has a much harder problem than the evidentiality model $\mathcal{M}$ from the previous section: $\mathcal{M}$ has access to the gold answer $\hat{y}$, while $\mathcal{G}^+$ does not.  Intuitively, we can get reasonably accurate evidentiality labels from $\mathcal{M}$ using the gold answer, then force $\mathcal{G}^+$ to predict those labels without access to the gold answer, in order to teach the encoder of $\mathcal{G}^+$ to better determine the relationship between $x$ and $p_i$.

\paragraph{Multi-task training. }
\label{sec:loss}
We conduct multi-task training of generation and evidentiality prediction. 
In particular, our framework minimizes a multi-task objective below:
\begin{equation}\label{eq:overall_multi_task}
    \mathcal{L} = \mathcal{L}_{gen}+ \lambda \mathcal{L}_{class}, 
\end{equation}
where $\lambda$ is a weighting parameter to balance the two objectives and would be tuned.
In \Eref{eq:overall_multi_task}, $\mathcal{L}_{gen}$ is formulated as follows:
\begin{equation}
    \mathcal{L}_{gen} = - \sum_{j}^T{\log {p(\hat{y}_j | y_{\small < j}, q, \tilde{\mathbf{P}}) }}, 
\end{equation}
where $\hat{y}_j$ denotes the $j$th token of the annotated gold answer $\hat{y}$. 
{Similarly}, evidentiality prediction objective $\mathcal{L}_{class}$ can be written as follows:
\begin{equation}\label{eq:class_loss}
    \mathcal{L}_{class} = - \sum_{i}^N{\log { p(e^{silver}_i | q, p_i )}}.
\end{equation}
Note that this probability is computed by a T5 decoder as a common practice~\cite{JMLR:v21:20-074}; even though $e^{silver}_i \in \{positive, negative\}$, the probability is normalized over T5's entire output vocabulary.\footnote{{We also tried to fine-tune a simple binary classification model using additional output layer on the top of T5 encoder. We found that this model performs much worse than T5-decoder-based classification model. }}
\begin{table*}[t!]
\small
    \centering
    \begin{tabular}{l| ccc | c|c }
\toprule
Dataset \& Task &  \multicolumn{3}{c}{\# of examples} & evaluation & Top-20 recall \\ 
& train & dev & test & metric &  (\%) \\ 
\midrule
{\bf 1. Open-domain QA} & &&&&\\
Natural Questions Open~\cite{kwiatkowski2019natural} & 79,168 & 8,757  & 3,610 & EM & 82.1 \\
TriviaQA unfiltered~\cite{JoshiTriviaQA2017} &  78,785 &  8,837 & 11,313 & EM & 75.2  \\\hline
\rule{0pt}{2ex} 
{\bf 2. Fact Verification} & &&&&\\
FEVER~\cite{thorne-etal-2018-fever} & 104,966  & 10,444 & 10,100 & Accuracy &  98.1 \\
FaVIQ-Ambig (A)~\cite{park2021faviq} & 17,008  & 4,260  & 4,688  & Accuracy & 100.0 \\\hline
\rule{0pt}{2ex} 
{\bf 3. Knowledge-enhanced Dialogue} & & & & &\\
Wizard of Wikipedia~\cite{dinan2018wizard} & 63,734 & 3,054&  2,944 & F1 & 96.2 \\
 \bottomrule
 \end{tabular}
    \caption{Dataset statistics. We experiment with three diverse knowledge-intensive NLP tasks across six datasets. ``Top 20 recall'' calculates if any of the top 20 passages include the answer strings (for open-domain QA datasets and FaVIQ-A) or comes from the provenance article (for FEVER and Wizard of Wikipedia) in the development set.  
    FEVER and Wizard of Wikipedia are based on the KILT~\cite{petroni-etal-2021-kilt} version.
    }
    \label{tab:data_stat}
\end{table*}

\section{Experimental Setups}
We experiment on three knowledge-intensive tasks: open-domain QA, fact verification, and knowledge-enhanced dialogue. 
Statistics for each dataset are provided in \Tref{tab:data_stat}. 

\subsection{Tasks, Datasets, and Metrics}
\paragraph{Open-domain QA.}
We use Natural Questions Open~\cite{kwiatkowski2019natural} and TriviaQA-unfiltered~\cite{JoshiTriviaQA2017} to evaluate our method on  open-domain QA. 
Natural Questions {consists of questions, long answers (e.g., gold evidence passages) and short answers (e.g., spans in the long answers), and the open-domain QA version is created by discarding questions that only have long answers or short answers whose length is longer than five tokens~\cite{lee-chang-toutanova:2019:ACL2019}.
} 
TriviaQA-unfiltered~\cite{JoshiTriviaQA2017} includes unfiltered 110K Trivia question and answer pairs. 
For both of the datasets, we use publicly available DPR retrieval results for training and inference data,\footnote{\url{github.com/facebookresearch/DPR}} and do not further fine-tune retrievers. 
Only the Natural Questions dataset has gold passage annotations {and we use the gold passage annotations to train the evidentiality labeling model $\mathcal{M}$ only.}
Following prior work~\cite{lee-chang-toutanova:2019:ACL2019}, we use Exact Match (EM) as our primary metric. 

\paragraph{Fact verification.} 
{
We use FaVIQ Ambig (FaVIQ-A; \citealt{park2021faviq}) and FEVER~\cite{thorne-etal-2018-fever} via the KILT benchmark~\cite{petroni-etal-2021-kilt} to evaluate our method on fact verification. 
}
FaVIQ-A is created from an information-seeking QA dataset, AmbigQA~\cite{min2020ambigqa} to pose realistic fact verification queries.
We use the baseline code provided by the authors of the FaVIQ dataset and KILT. 
We use accuracy as our evaluation metric. 

\paragraph{Knowledge-enhanced dialogue.} 
{
We use Wizard of Wikipedia~(WoW; \citealt{dinan2018wizard}) to evaluate our method on knowledge-enhanced dialogue.
}
We use the officially available KILT DPR baseline codes~\cite{petroni-etal-2021-kilt}\footnote{\url{github.com/facebookresearch/KILT/blob/main/kilt/retrievers/README.md}} to obtain passages and evaluate downstream F1 score. 

\begin{table*}[t!]
    \centering
    \begin{subtable}[b]{0.35\textwidth}
    \centering
    \resizebox{0.95\columnwidth}{!}{
    \begin{tabular}{l|c c| cc }
\toprule
Models & \multicolumn{2}{c}{NQ EM} &  \multicolumn{2}{c}{TQA EM} \\ 
 & dev & test & dev & test \\
\midrule
RAG (large) & -- & 44.5 & -- & 56.8  \\
FiD (base) & 46.9 & 48.3 & 67.1 & 67.2 \\\hline
Ours(base) & 47.8  &  49.8 & 67.7 & 67.8 \\ \hline \hline 
R2D2 (large*) & -- & 55.0 & -- & 69.9 \\
 \bottomrule
 \end{tabular}
    }
    \caption{} 
    \label{tab:main_results_openqa}  
    \end{subtable} \hfill
    \begin{subtable}[b]{0.35\textwidth}
    \centering
    \resizebox{0.95\columnwidth}{!}{
    \begin{tabular}{l| c c| cc}
\toprule
Models &  \multicolumn{2}{c}{FaVIQ-A} & \multicolumn{2}{c}{FEVER} \\ 
 & dev & test &  dev & test  \\
\midrule
DPR+BART (large) & 66.9 & 64.9  &  88.1 &  86.7\\
DPR+BART (base) & 60.2 & -- &  --  & --  \\
RAG (large) &  -- & -- & 87.7 & 86.3 \\
FiD (base)  &  67.8 & 64.3  & 89.5 & -- \\\hline
Ours (base) &   {\bf 69.6}  &   {\bf 65.7} & {\bf 89.8} & {\bf 88.5} \\
 \bottomrule
 \end{tabular}
    }
    \caption{} 
    \label{tab:faviq}
    \end{subtable}
        \begin{subtable}[b]{0.28\textwidth}
    \centering
    \resizebox{0.95\columnwidth}{!}{
    \begin{tabular}{l|c c}
\toprule
& \multicolumn{2}{c}{WoW} \\
Models & dev & test  \\ 
\midrule
DPR+BART (large) &15.5 &  15.2  \\
RAG (large) & 13.8 & 13.1 \\
FiD (base) &  16.9 & --  \\\hline
Ours (base) & {\bf 17.9} & {\bf 17.3} \\
 \bottomrule
 \end{tabular}
    }
    \caption{}
    \label{tab:wow}
    \end{subtable}
        \caption{{\bf Main Results}. ``base'' and ``large'' denote the base generator model sizes (e.g., T5-large, BART-base). (a) Performance on Natural Questions Open and TriviaQA unfiltered. ``NQ'' denotes Natural Questions Open, ``TQA'' denotes TriviaQA unfiltered. The state-of-the-art model is R2D2 from \citet{fajcik-etal-2021-r2-d2}, which has 1.29 billion parameters (more than twice more parameters than our model), consisting of a ranker and two reader models with ELECTRA~\cite{Clark2020ELECTRA}-large and T5-large. 
        (b) Performance on FaVIQ-A and FEVER. 
        Previous best model is DPR+BART (large) from \citet{park2021faviq} and \citet{petroni-etal-2021-kilt} on FaVIQ-A and FEVER, respectively. 
        (c) Performance on Wizard of Wikipedia (WoW). 
        The best published model on the development set is DPR+BART (large) from \citet{petroni-etal-2021-kilt}. {Test set results of WoW and FeVER are based on the leaderboard results at the time of the paper submission (January, 2022). }
        }
\end{table*}
\subsection{Baselines}
We use FiD~\cite{izacard2020leveraging} as our primary baseline using their official implementation.\footnote{\url{github.com/facebookresearch/FiD}}
In addition, we report results from the best {published, }publicly available generator models for each dataset including RAG~\cite{lewis-etal-2020-bart} and DPR + BART~\cite{petroni-etal-2021-kilt}.  
{For FEVER and WoW, we also compare our method with the published models on the KILT leaderboard.\footnote{\url{ai.facebook.com/tools/kilt/}}}

\subsection{Hyper parameters}
Due to the computational budget, we use T5's base-size models throughout our experiments {for our evidentiality-guided generator. 
For our evidentiality labeling model $\mathcal{M}$, we use a RoBERTa~\cite{liu2019roberta}-base binary classification model.}
If not specified, we use the top 20 passages during training and inference, which also reduces the computational times from the original FiD model that uses top {100} passages. 
We train the models for 120k steps using 8 GPUs with 24 GB memory and take the checkpoint that achieves the highest score on the development set. 
The batch size is set to 1 and to imitate the larger batch size, we set the gradient accumulation step to be 4. 
The learning rate is set to $10^{-5}$ and the number of warm-up steps is 1000. 
We set $\lambda$ to be 0.5 for open-domain QA and dialogue, and 0.1 for fact verification. 
See more details in Appendix. 
\section{Results and Analysis}
Our approach significantly improves over its direct counterpart on all datasets, and outperforms all prior published results on FaVIQ-A, FEVER and WoW, advancing their state-of-the-art performance.
\subsection{Task Results}
\paragraph{Open-domain QA.}
\Tref{tab:main_results_openqa} shows experimental results on the two open-domain QA datasets. 
On Natural Questions Open, we improve the performance over FiD by 1.5 EM score. 
We observe performance improvements over FiD on TriviaQA as well. 
{It should be noted that on open-domain QA, most of the recent models \cite[e.g.,][]{fajcik-etal-2021-r2-d2} contain a few times more parameters than our model or use improved retrievers \cite{izacard2020distilling}, both of which are beyond our computational budgets.  
Our results represent state-of-the-art performance for models with access to similar computational resources, and our contributions should be complementary to work focusing on improving retrieval components. }

\paragraph{Fact verification.}
\Tref{tab:faviq} shows the experimental results on FaVIQ-A and FEVER. 
In addition to the original paper's baseline, we have fine-tuned a BART-base baseline using their original public codebase (DPR+BART (base)) for a fair comparison.\footnote{\url{github.com/faviq/faviq}}
Our model significantly outperforms the prior best model, DPR+BART (large), on FaVIQ-A by a large margin. 
Our model also significantly outperforms FiD on FaVIQ by 1.8\% on the development set and 1.4\% on the test set, yielding state-of-the-art performance on this dataset. 
Our evidentiality-guided generator also outperforms other models on FEVER.
On the FEVER hidden test set,\footnote{\url{eval.ai/web/challenges/challenge-page/689/leaderboard/1899}} our model yields 88.5\% down-stream accuracy and ranks second among all submissions, outperforming all of prior published work~\cite{maillard-etal-2021-multi,petroni-etal-2021-kilt,lewis-etal-2020-bart}.

\paragraph{Knowledge-enhanced dialogue.}
\Tref{tab:wow} shows the experimental results on the Wizard of Wikipedia dataset.
Our model outperforms prior work using larger base models and improves the F1 score from the base FiD model by 1.0. 
On the test set,\footnote{\url{eval.ai/web/challenges/challenge-page/689/leaderboard/1909}} our model yields 17.3 F1, outperforms all other published work and ranks fourth among all submissions (the top three are unpublished). 

\subsection{Analysis}
\subsubsection{Ablation Study}
We study the impact of different components of our method by comparing the full method with other variants. 

\noindent {\bf - Multi-task} does not use our multi-task objective and only trains with $\mathcal{L}_{gen}$, which is theoretically equivalent to FiD.  

\noindent {\bf - $\mathbf{E}^{silver}$ mining} uses the multi-task training but does not use our method to find evidentiality silver labels. 
Instead, it relies on task-specific heuristics (e.g. string match) that have been used by prior work \cite{chen2017reading}. For WoW and FaVIQ-A, where we cannot locate gold answers in the retrieved context to label evidentiality, we use additional meta-data such as gold Wikipedia article titles available in the original datasets~\cite{petroni-etal-2021-kilt}.  
It should be noted that that additional metadata is often unavailable in most of the datasets, and this variant for WoW and FaVIQ can be considered as a ground-truth setting. See more details in Appendix. {Moreover, relying on this dataset-specific metadata limits models’ applicability to wider NLP datasets and tasks.}  {Note that our method does not use this additional metadata, so this variant can get higher numbers than our model.}


\noindent {\bf - LOO-gen.} uses the multi-task training but removes our leave-one-out-generation strategy for collecting evidentiality labels. It only incorporates the first step of training the evidentiality model over Natural Questions only. 

\begin{table}[t!]
\small
\addtolength{\tabcolsep}{-1.5pt}
    \centering
    \begin{tabular}{l|c c c }
\toprule
Models & NQ & FaVIQ-A & WoW \\ 
Metric & EM & Acc &  F1 \\ 
\midrule
Ours &  {\bf 47.9} &  {\bf 69.6} & 17.9  \\\hdashline
- multi-task  & 46.9 &  67.8  & 16.9   \\
- $\mathbf{E}^{silver}$ mining  & 47.3 &  69.1* & {\bf 18.0*} \\
- LOO-gen. &  47.6 &  69.2  &  17.7 \\
 \bottomrule
 \end{tabular}
    \caption{Ablation results. All results are based on the performance on development set of the three datasets. ``NQ'' denotes Natural Questions Open and ``WoW'' denotes Wizard of Wikipedia. $^*$ in the FaVIQ-A and WoW columns indicate that a model is trained with additional metadata our evidentiality-guided generator does not use during training. 
    }
    \label{tab:ablatons}
\end{table}

\Tref{tab:ablatons} reports the ablation results. 
There is a clear drop  when removing the multi-task auxiliary learning, especially on FaVIQ-A, where a model needs to precisely assess the evidence and reason, without being distracted by a simple lexical overlap~\cite{park2021faviq}. 
Removing {\it $\mathbf{E}^{silver}$} mining drops the performance on all of the three datasets, indicating the effect of mining evidentiality labels, instead of relying on string matching heuristics. 
{Note that especially on FaVIQ-A or WoW, this ``-{\it $\mathbf{E}^{silver}$} mining'' uses oracle gold annotations, which are not used by ours. 
By removing the necessity of having access to task-specific heuristics or those additional annotation, our method is easily applicable to a task or a new dataset. }
{
Finally, the performance drop when removing LOO-gen. shows the impact of our leave-one-out approach in collecting evidentiality labels for target tasks to train $\mathcal{M}$.
}

\subsubsection{Evaluating Evidentiality Labels}
\Tref{table:human_analysis_nq} shows human analysis over evidentiality positive and negative labels obtained by our method over randomly selected samples. In particular, we randomly sample 50 Natural Questions development questions and sample 2 positive passages and 2 negative passages  (if applicable) with answer strings for each question. The authors manually analyze (i) if the positive passages actually provide sufficient evidence to answer, and (ii) if the negative passages actually {\it do not} provide sufficient evidence to answer, despite the existence of the gold answer strings. 
We found that in 95\% of the mined positive passages provide sufficient evidence to answer, while only 4\% of the negative passages do not; in other words, the predictions are correct 95\% of the positive passages and 96\% of the negative passages.  

\begin{table}[t!]
    \centering
    \begin{subtable}[b]{0.40\textwidth}
    \centering
    \resizebox{0.9\columnwidth}{!}{
\begin{tabular}{cc| c }
\toprule
$e^{silver}$ & $\hat{e}$ & \% \\ 
\midrule
\texttt{pos} & \texttt{pos}  & 95 \\
\texttt{pos}   & \texttt{neg} & 5 \\ \hdashline
\texttt{neg} & \texttt{pos}   &  4 \\
\texttt{neg} & \texttt{neg} & 96 \\
\bottomrule
\end{tabular}
    }
    \caption{} 
    \label{table:human_analysis_nq}  
    \end{subtable} \hfill
    \begin{subtable}[b]{0.55\textwidth}
    \centering
    \resizebox{0.9\columnwidth}{!}{
\begin{tabular}{l|c }
\toprule
(category) relevance & \% \\
\midrule
(1) $p_\mathcal{G}^+ > p_\mathcal{G}$ &  43  \\
(2) $p_\mathcal{G}^+ < p_\mathcal{G}$  & 14 \\
(3) $p_\mathcal{G}^+ = p_\mathcal{G} = 0 $  & 29 \\
(4) $p_\mathcal{G}^+ = p_\mathcal{G} = 1$  & 14 \\
\bottomrule
\end{tabular}
    }
    \caption{} 
    \label{table:human_analysis}
    \end{subtable}
        \caption{(a) Human analysis over evidentiality positive and negative labels obtained by our method. $e^{silver}$ denotes predictions made by $\mathcal{M}$ while $\hat{e}$ denotes the evidentiality  labeled by human annotators. \texttt{pos} denotes evidentiality-positive while \texttt{neg} denotes evidentiality negative. (b) Qualitative evaluation of $\mathcal{G}$ and $\mathcal{G}^+$. $p_\mathcal{G}$ and $p_\mathcal{G}^+$ denotes the relevance between the input and the passages most attended by $\mathcal{G}$ and $\mathcal{G}^+$, respectively. }  
\end{table}


\subsubsection{Qualitative evaluation of $\mathcal{G}$ and $\mathcal{G}^+$} 
We conduct a systematic qualitative analysis on the FaVIQ-A predictions made by a base generator $\mathcal{G}$ and our evidentiality-guided generator $\mathcal{G}^+$. 
{
We study the claims in the evaluation set that $\mathcal{G}$ and $\mathcal{G}^+$ provide different prediction classes (793 out of the total 4,260 claims). 
We observe $\mathcal{G}^+$ provides the correct labels in 54\% of these cases. 
We further filter out the cases where the two models provide the highest attention scores to similar passages, leading to 192 claims. }
The authors of this paper manually inspect all of those 192 claims and classify them into four categories: (1) $\mathcal{G}^+$ attends to a more relevant passage ($p_\mathcal{G}^+ > p_\mathcal{G}$), (2) $\mathcal{G}$ attends to a more relevant passage ($p_\mathcal{G}^+ < p_\mathcal{G}$ ), (3) the models attend to equally-irrelevant passages ($p_\mathcal{G}^+ = p_\mathcal{G} = 0$), (4) both of them attend to equally-relevant passages ($p_\mathcal{G}^+ = p_\mathcal{G} = 1$). 
The \Tref{table:human_analysis} (b) results show that $\mathcal{G}^+$ attends to the passages that are more relevant to the input claims. 
{After further inspection, we found that $\mathcal{G}$ sometimes generates the right class, even if it gives the highest attention to a less relevant passage, explaining a smaller accuracy gap between the two models. }
This probably happens due to the nature of the task (e.g., two-way classification). 
We show some examples in \Tref{tab:qual_faviq} in the Appendix.

\subsubsection{Performance on Hard Subsets}{
We automatically collect challenging instances from FaVIQ-A and Trivia QA development set, to see if there is an even more notable gap between $\mathcal{G}$ and $\mathcal{G}^+$ on those harder examples.}
To this end, we feed the top one retrieved passages with the input queries to the two generators and label questions that both models can answer correctly given top passages only {\it easy}, otherwise {\it hard}. 

\Tref{tab:analysis_topone} shows the models' performance on the easy and hard subsets. 
In FaVIQ-A, the performance gap between two models on the harder subset is larger than the gap on the easy subset (i.e., 1.7 \% v.s. 1.1\% accuracy gap). 
Interestingly on FaVIQ-A, both models show somewhat low performance on the easy subset, where two models originally succeed to answer correctly given a single passage only. 
This is probably because the models are distracted by other passages when questions are actually simple and can be answered by top passages. 
On the other hand, the full accuracy of these top one passage only-variants is low (Ours: 54.7 \% accuracy, FiD: 53.4\%), suggesting the effectiveness of reading more passages. 
On the TriviaQA easy subset, both models show nearly 95\% EM, showing little performance gap between the two models, while there is a notable performance gap between the two models on the hard subset. 
{
These results indicate that our method is more effective on harder examples that require carefully assessing and reasoning the passages beyond the top one. 
}

\begin{table}[t!]
\small
\addtolength{\tabcolsep}{-1.5pt}
    \centering
    \begin{tabular}{l|c c | cc }
\toprule
dataset & \multicolumn{2}{c}{FaVIQ-A (Acc.)} & \multicolumn{2}{c}{TQA (EM)} \\ \hline
split(\#) & easy(1.7k) & hard(2.5k)& easy(4.0k) & hard(8.8k)\\ 
\midrule
FiD &  74.5 & 62.9   &  94.8 & 37.1 \\
Ours & 75.6 & 64.6  & 94.8  &  36.0 \\
 \bottomrule
 \end{tabular}
    \caption{ 
    Performance on {\it easy} and {\it hard} subsets of FaVIQ-A and TriviaQA (TQA), decided by top one only models' predictions. 
    The numbers inside parenthesis show the number of the examples included in the easy and hard subsets. 
    }
    \label{tab:analysis_topone}
\end{table}
\section{Related Work}

\paragraph{Retrieval-augmented generation.}
Retrieval-augmented generators leverage retrievers such as Dense Passage Retriever~\cite{karpukhin2020dense} or BM25~\cite{robertson2009probabilistic} to find evidence from many passages, and feed those retrieved passages with the original query to competitive pre-trained generators such as BART~\cite{lewis-etal-2020-bart} and T5~\cite{brown2020language}. 
They achieve competitive performance across different knowledge-intensive NLP tasks~\cite{izacard2020leveraging,glass2021zero,paranjape2021hindsight,park2021faviq,borgeaud2021improving}. 
Recent work improves the retrieval component~\cite{paranjape2021hindsight,maillard-etal-2021-multi} or introduces another passage re-ranking modules~\cite{fajcik-etal-2021-r2-d2} for further improvements. Our work focuses on improving the generator component, which has been underexplored in the literature. Our work is complementary to those prior work focusing on improving the retrieval components of retrieval-augmented generation. 

\paragraph{Unsupervised evidence selection for multi-hop QA.}
Recently, \citet{lee-etal-2021-robustifying} introduce evidentiality-guided training for multi-hop question answering such as HotpotQA~\cite{yang-etal-2018-hotpotqa}{, which mines evidence sentences by adding or removing them to create counterfactual cases, and train a QA model with a regularization term to avoid overconfidence on negative passages. 
{
Recent work~\cite{nishida2019qfe,fajcik-etal-2021-r2-d2} introduces multi-task learning of answer generation and evidence selection in the area of multi-hop QA or open-domain QA, but these approaches often rely on evidence annotations or heuristics (e.g., answer string matching) for supervising multi-task loss, which is unavailable in most of the datasets and tasks such as knowledge-enhanced dialogue or fact verification.}
Several prior work attempts to learn to find evidence sentences in unsupervised manners in multi-hop QA~\cite{chen2019multi,yadav2019quick,perez-etal-2020-unsupervised}, whereas our work uses evidentiality to improve the generator components via multi-task training for diverse knowledge-intensive tasks, going beyond QA alone. }

\paragraph{Entailment-based approaches to improve QA.}
{Assessing evidentiality of a passage given a question and a final output can be framed as an entailment task. }
Using entailment models to enhance the performance of QA tasks has been extensively studied~\cite{harabagiu-hickl-2006-methods,sacaleanu-etal-2008-entailment,abacha2019question,trivedi-etal-2019-repurposing}.
\citet{iyer-etal-2021-reconsider} introduce an NLI-based reranker to improve open-domain QA performance, and 
\citet{chen2021can} use NLI models to calibrate the answer reliability. 
They focus on improving the final answers, while we incorporate evidentiality more directly into the base model. 
\section{Conclusion}
Augmenting pre-trained generation models with retrievers has shown to be effective in many knowledge-intensive tasks; however, they often rely on spurious cues or generate hallucinations during inference. 
We introduce a multi-task learning objective the combines answer generation and evidentiality prediction. We propose task-agnostic data mining techniques to obtain {\it silver} evidentiality labels to enable this auxiliary training. Our experiments across five datasets show large performance improvements over baselines and our evidentiality-guided generator advances the state-of-the-art performance on FaVIQ-Ambig, FEVER and WoW. 
Our analysis shows that multi-task learning and silver evidentiality mining both contribute to the performance improvements by helping the model learn to focus on and generate answers from more relevant passages.

\section*{Broader Impact and Ethical Implications}
{
Retrieval-augmented generation models have shown state-of-the-art performance in a range of knowledge-intensive NLP tasks such as QA, fact verification, dialogue and long-form QA. 
However, prior work found that they often hallucinate~\cite{xu2021attention} or are easily distracted by irrelevant evidence~\cite{longpre2021entitybased}. 
Those issue can cause serious risks especially when those technologies are applied to certain domains such as health care or politics.    
This work aims at solving those challenges and experimental results show that our proposed approach improves the performance in diverse downstream applications, learning to focus on more relevant passages than the original baseline. 
Although our model can still cause generation errors, our evidentiality predictor now provides predictions of evidentiality labels, which help practitioners understand the models' behavior. 
We have released our code and trained models so that follow-up work can reproduce and improve our method. 
}

\section*{Acknowledgements}
This research was supported by NSF IIS-2044660, ONR N00014-18-1-2826,
the Allen Distinguished Investigator Award, the Sloan Fellowship, and the Nakajima Foundation
Fellowship. 
We thank the anonymous reviewers, the members of the UW NLP group and Allen NLP for their insightful discussion and feedback on this paper. 

\bibliography{anthology}
\bibliographystyle{acl_natbib}
\clearpage
\appendix

\section*{Appendix}
\label{sec:appendix}
\section{Preliminary Experiments and Analysis}
\subsection{Analysis on a Base Generator $\mathcal{G}$}
\label{sec:fid_analysis}

\paragraph{Error analysis.}
We conduct a detailed error analysis on the base generator, FiD. 
We manually analyzed 50 errors in the Natural Questions development set to understand what causes the errors. 
Although 23 errors are due to the annotation errors (e.g., correct answer aliases are not covered by the original data; questions are highly ambiguous as pointed by \citealt{min2020ambigqa,asai2020unanswerable}), we found that the model often succeeds in retrieving the right evidence but fails to generate the answers based on the passages with supporting evidence. 
We show the top attended passages for sampled questions in \Tref{tab:attended_passages_analysis_error}. 
Although those passages have high lexical overlap with the questions, they are often irrelevant or about the different entities in the same genre (e.g., last name, movie). Yet, during training, the model is only given the final output supervision signal, making it difficult to distinguish the passages with sufficient evidence to answer from the ones without evidence. 

\paragraph{Memorization issues.}
We also found that when the retrieved passages are not evident the model more often generates incorrect answers memorized during training, without carefully accessing the context. 
In the questions where FiD fails to generate the correct answers, more than 5\% of the answers are not sub-spans of any of the retrieved passages, while in the questions FiD succeeds to answer 99.5\% of the answers are copied from the passages. 
Moreover, in the success cases, the predicted answers are the sub-spans of the top 10 passages in 96\% of the cases, while in the error cases, only 79\% of the predicted answers are copied from the top 10 passages. 
Those findings are consistent with the ones observed by \citet{xu2021attention}.
Recently, \citet{longpre2021entitybased} found that the generative QA models often generate the answers memorized during fine-tuning, when they observe more unreliable passages during training. 

\subsection{Evidentiality Negative Passages among Top Retrieved Passages }
We manually analyze 20 sampled Natural Questions training questions where at least of one of the top 3 passages retrieved by DPR include the annotated gold answers, to see if including answer strings entails evidentiality. Labeling passages with answer strings positive have been commonly used in open-domain QA~\cite{chen2017reading,karpukhin2020dense}, but prior work found that those passages are often spurious~\cite{min2019discrete}.
We found that in 30\% of the cases, the passages with answer strings do not actually provide evidence to answer the input questions. 
We shows the examples in \Tref{tab:evidentiality_analysis}. 
Training a model with distantly supervised approaches have been widely used in open-domain QA, but particularly in the current retrieved-augmented training schema, this approach can cause huge learning noises. 
It also should be noted that those passages are all from top 3 retrieved results, which are expected to be highly related to the input queries. 

\begin{table*}[ht!]
\center
\begin{tabular}{p{0.95\linewidth}}
\toprule 
{\bf Q}: who played \underline{mary} in christmas with the kranks \\
{\bf A}: Felicity Huffman
\\\hline
\multirow{5}{\linewidth}{{\bf Christmas with the Kranks}:~~Christmas with the Kranks Christmas with the Kranks is a 2004 American Christmas comedy film based on the 2001 novel ``Skipping Christmas'' by John Grisham. It was directed by Joe Roth and written and produced by Chris Columbus. It stars Tim Allen and Jamie Lee Curtis as a couple who decide to skip Christmas one year since their daughter is away, much to the chagrin of their neighbors. .}\\
\\
\\
\\
\\\bottomrule
\\
\toprule 
{\bf Q}: \underline{hyori bed and breakfast} season 2 air date  \\
{\bf A}: February 4, 2018
\\\hline
\multirow{5}{\linewidth}{\underline{\bf Queen Sugar}:~~On March 11, 2016, it was announced that Marycarmen Lopez also was cast as regular. On August 1, 2016, the series was renewed for a second season ahead of its television premiere which aired in a two-night premiere on June 20 and June 21, 2017. The second season premiered on OWN in a two episode special on June 20 and 21, 2017. 
The show was renewed for a third season on July 26, 2017. The third season premiered in a two-night special on May 29 and May 30, 2018. On August 8, 2018, the series.}\\
\\
\\
\\
\\
\\\bottomrule
\\
\toprule 
{\bf Q}: where does the last name \underline{waters} come from \\
{\bf A}: Wales and Yorkshire
\\\hline
\multirow{5}{\linewidth}{ {\bf Bywater (surname)}:~~\underline{Bywater} (surname) Bywater is an uncommon English surname of Anglo-Saxon origin and can most frequently be found in the English region of Yorkshire. It is a topographical surname given to those who were situated near a body of water. Bywater is an uncommon surname of Anglo-Saxon origin. The name derives from the merger of the Old English words ``bi'' and ``waeter'' to form ``biwaeter''. Topographical surnames are among the earliest created, because natural and artificial features in the.}\\
\\
\\
\\
\\
\\\bottomrule
\\
\toprule 
{\bf Q}: who was last person to \underline{be executed in us} \\
{\bf A}: Ruben Cardenas Ramirez
\\\hline
\multirow{4}{\linewidth}{ {\bf Billy Bailey}:~~He became only the third person to be hanged in the United States since 1965 (the previous two were Charles Rodman Campbell and Westley Allan Dodd, both in Washington) and the first person hanged in Delaware in 50 years. As of 2018, he remains the last person to be \underline{executed by hanging in the United States}.}\\
\\
\\
\\\bottomrule
\\
\toprule 
{\bf Q}: what is the largest ethnic group \underline{in mexico today} \\
{\bf A}: K'iche'
\\\hline
\multirow{3}{\linewidth}{ {\bf Mexican-American middle class}:~~the Latino/a population of \underline{the United States} is the nation’s largest racial/ethnic minority group, constituting 17.6 percent of the total population. At two thirds of the Latino/a ethnic category, Mexicans are by far the largest national origin group. .}\\
\\
\\\bottomrule
\\
\end{tabular}
\caption{Examples of the top attended spurious passages in the questions where the base generator $\mathcal{G}$ failed to generate the correct answers. The underlined phrases contradict the input queries, while those passages generally have high lexical overlap with the given input queries.}
\label{tab:attended_passages_analysis_error}
\end{table*}

\begin{table*}[ht!]
\center
\begin{tabular}{p{0.95\linewidth}}
\toprule 
{\bf Q}: who is in charge of enforcing the pendleton act of 1883 \\
{\bf A}: United States Civil Service Commission
\\\hline
\multirow{3}{\linewidth}{ {\bf 1. Pendleton Civil Service Reform Act}:~~Pendleton Civil Service Reform Act The Pendleton Civil Service Reform Act (ch. 27, ) is a United States federal law enacted in 1883 that mandated that positions within the federal government should be awarded on the basis of merit.}\\
\\
\\\hdashline
\multirow{2}{\linewidth}{ {\bf 2. United States Civil Service Commission}:~~The Pendleton law was passed in part due to public outcry over the assassination of President Garfield.}\\
\\\hdashline
\multirow{3}{\linewidth}{{\bf 3. Pendleton Civil Service Reform Act}:~~The Act was written by Dorman Bridgman Eaton, a staunch opponent of the patronage system who was later first chairman of the {\bf United States Civil Service Commission}.}
\\
\\
\\\bottomrule
\\
\toprule 
{\bf Q}: who plays skyler on lab rats elite force \\
{\bf A}: Paris Berelc
\\\hline
\multirow{2}{\linewidth}{ {\bf 1. Lab Rats: Elite Force}:~~The series is a combined spinoff of ``Lab Rats'' and ``Mighty Med'' and stars William Brent, Bradley Steven Perry, Jake Short, {\bf Paris Berelc}, and Kelli Berglund. }\\
\\\hdashline
\multirow{3}{\linewidth}{ {\bf 2. Lab Rats: Elite Force}:~~ Elite Force is an American comedy television series created by Chris Peterson and Bryan Moore that aired on Disney XD from March 2 to October 22, 2016. ... stars William Brent, Bradley Steven Perry, Jake Short, {\bf Paris Berelc}, and Kelli Berglund. }\\
\\
\\\hdashline
\multirow{4}{\linewidth}{{\bf 3. Lab Rats: Elite Force}:~~On September 3, 2015, it was announced that ``Lab Rats'' and ``Mighty Med'' would have a joint spinoff series called ``Lab Rats: Elite Force''. Only William Brent, formerly credited as Billy Unger, and Kelli Berglund from ``Lab Rats'' and Bradley Steven Perry, Jake Short, and {\bf Paris Berelc} from ``Mighty Med'' were announced as returning for the new spinoff series. . }
\\
\\
\\
\\\bottomrule
\\
\toprule 
{\bf Q}: who developed the first periodic table with 8 columns \\
{\bf A}: Dmitri Mendeleev 
\\\hline
\multirow{3}{\linewidth}{ {\bf 1. Periodic table}:~~In 1923, Deming, an American chemist, published short (Mendeleev style) and medium (18-column) form periodic tables. Merck and Company prepared a handout form of Deming's 18-column medium table, in 1928, which was widely circulated in American schools. }\\
\\
\\\hdashline
\multirow{3}{\linewidth}{ {\bf 2. History of the periodic table}:~~their decision by saying that such ``'theoretical'' topics might be controversial. The importance of Newlands' analysis was eventually recognised by the Chemistry Society with a Gold Medal five years after they recognised Mendeleev's work.}\\
\\
\\\hdashline
\multirow{4}{\linewidth}{{\bf 3. History of the periodic table}:~~the work of {\bf Dmitri Mendeleev} had been published. In 1864, the English chemist John Newlands classified the sixty-two known elements into eight groups, based on their physical properties. Newlands noted that many pairs of similar elements existed, which differed by some multiple of eight in mass number, and was the first to assign them an atomic number. }
\\
\\
\\
\\\bottomrule
\end{tabular}
\caption{Examples of the top three passages retrieved by a trained $\mathcal{R}$ (DPR). {We make the phrases matching the gold answers bold in the retrieved passages. }}\label{tab:evidentiality_analysis}
\end{table*}

\begin{table*}[ht!]
\center
\begin{tabular}{p{0.95\linewidth}}
\\
\toprule 
{\bf Task}: Fact Verification \\\midrule
{\bf Claim$_1$}: jimmy perry had a cameo for the role of charlie cheeseman in dad's army. \\
{\bf Label$_1$}: SUPPORTS
\\\hdashline
\multirow{3}{\linewidth}{{\bf Jimmy Perry}:~~Despite the doubts, the first episode was screened on 31 July 1968, with \hl{Perry making a cameo appearance} as the entertainer Charlie Cheeseman in the sixth episode, "Shooting Pains".}\\\\\\\hline
{\bf Claim$_2$}: John Glenn was a military test pilot. \\
{\bf Label$_2$}: SUPPORTS
\\\hdashline
\multirow{2}{\linewidth}{{\bf John Glenn}:~~\hl{Glenn's first flight test assignment}, testing the FJ-3 Fury, nearly killed him when its cockpit depressurized and its oxygen system failed.}\\
\\\bottomrule
\\
\toprule 
 {\bf Task}: Knowledge-enhanced Dialogue \\\midrule
{\bf Contexts$_1$}: Purple is such a good color. \\
{\bf Response$_1$}: yep, its in between red and blue  \\\hdashline
\multirow{3}{\linewidth}{{\bf Purple}:~\hl{~Purple is a color intermediate between blue and red}. It is similar to violet, but unlike violet, which is a spectral color with its own wavelength on the visible spectrum of light, purple is a secondary color made by combining red and blue.}\\\\\\\hline
\multirow{5}{\linewidth}{{\bf Contexts$_2$}: I was a really good skateboarder when i was young, its an action sport which involves riding and performing tricks, have you used a skateboard ::: i tried wjhen i was younger but i failed horribly!haha ::: hah, yes its really hard,  first skateboards started with wooden boxes with wheels attached to the bottom, it was an invention from the people ::: i think i would have done alot better on a box with wheels! lol thats so cool. when was the first one invented? }\\\\\\\\\\
{\bf Response$_2$}: in the early 1900's it started, now there are 11.08 million active skateboarders in the world! \\\hdashline
\multirow{6}{\linewidth}{{\bf Electric skateboard}:~~An electric skateboard is a personal transporter based on a skateboard. The speed is controlled by a hand-held throttle or weight-shifting and the direction of travel is adjusted by tilting the board to one side or the other. The MotoBoard, which was gasoline-powered was released in the summer of 1975, but were banned in California due to their noise and pollution in 1997. Louie Finkle of Seal Beach, California is often cited as an originator of the modern electric skateboard, offering his first wireless electric skateboard .}\\\\\\\\\\
\\\bottomrule
\end{tabular}
\caption{{Examples of the positive examples newly mined by leave-one-out generation approach. ``:::'' in the contexts for the knowledge-enhanced dialogue example indicates the change of the speakers.} }
\label{tab:loo_paragraphs_results}
\end{table*}

\begin{figure*}[h!]
  \includegraphics[width=0.95\linewidth]{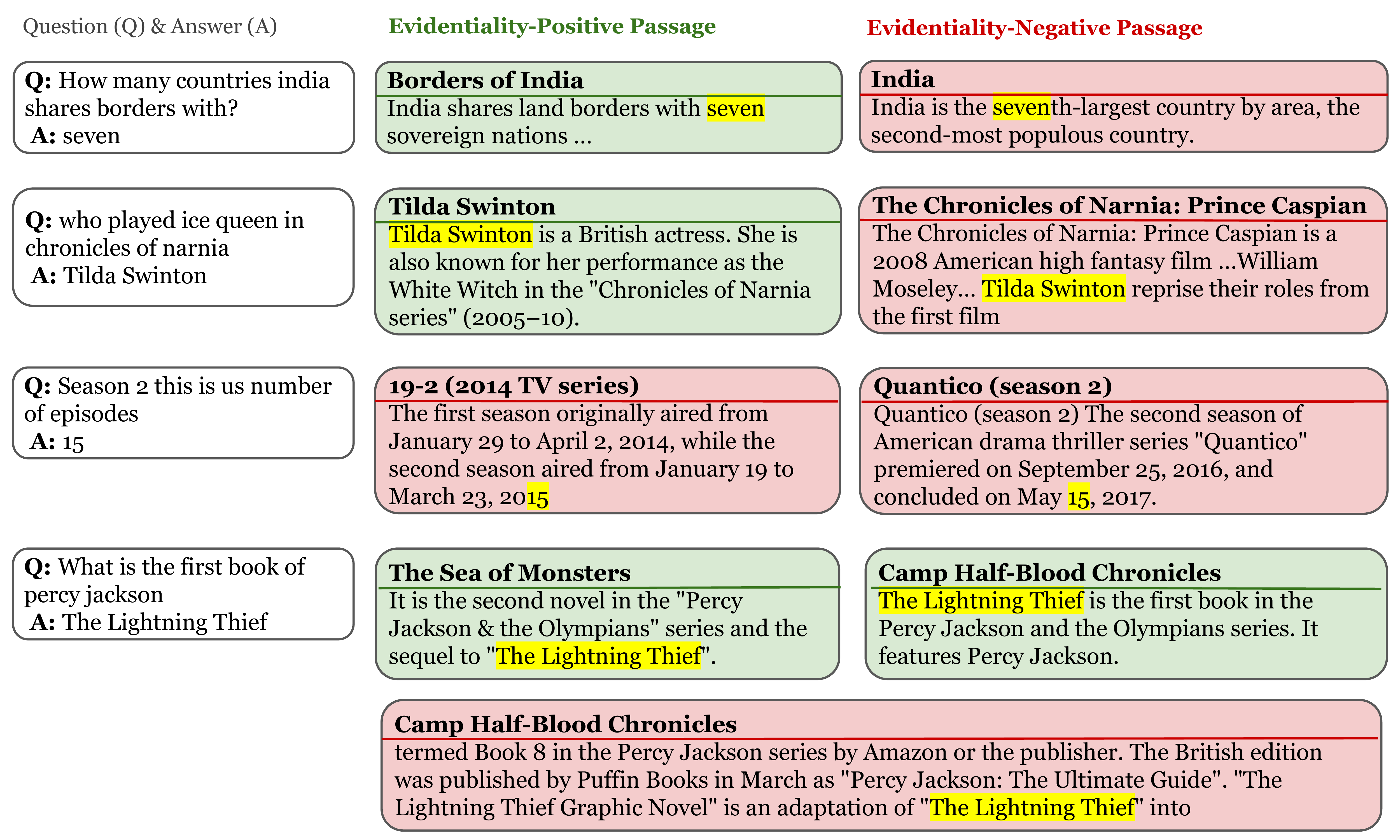}
  \caption{Examples of newly mined evidentiality examples for Natural Questions.}
  \label{img:new_evidentiality_example}
\end{figure*}

\section{Details about $\mathcal{M}$ and Resulting $\mathbf{E}^{silver}$}
\subsection{Lack of the {\it Gold} Evidentiality Labels}
Most datasets and tasks only include query-answer $(x, \hat{y})$ annotations and do not include evidentiality labels $\hat{\mathbf{E}}$ for passages $\mathbf{P}$. 
Some datasets with gold evidence annotation, such as Natural Questions, cover subsets of gold passages from certain Wikipedia articles, whereas $\mathcal{\mathbf{P}}$ possibly includes unlabeled gold passages from another article.
{
Where gold annotations are not available, a common heuristic is to use the answer string as distant supervision~\cite{mintz-etal-2009-distant}, labeling all passages that include the answer string as evidentiality positive.  This heuristic can create false-positive annotations---for instance, $p_2$ in \Fref{img:overview} includes the answer string ``seven'' but is irrelevant to the input query.
{Not only being noisy, this heuristic cannot be used for open-ended generation or answer classification}
such as knowledge-enhanced dialogue and fact verification. 
}
\subsection{Task-specific Details for Leave-one-out Generation}
\paragraph{Open-domain QA.}
To collect new positive and negative data using leave-one-out generation, we consider top 20 passages retrieved for all of the original training data queries, and then split 20 passages into two ten-passage chunks. 
We then run a trained FiD model for 10 times, masking $i$th passage at the $i$th iteration. 
We consider $i$th passage $p_i$ positive when and only when FiD fails to generate the correct answer when $i$th passage is masked. 
We also consider $p_i$ (hard-)negative when and only when FiD succeeds to answer correctly when $i$th passage is masked, as we assume that the $i$th passage can be highly distracting or confusing, misleading the generator.

\paragraph{Fact verification.}
As fact verification is a classification task, using the same methodology as open-domain QA may not be desirable---when we run a model ten times, it is likely to predict both correct and incorrect classes for multiple times, and we may not be able to mine the useful positive and negative passages. 
For the two fact verification datasets, we consider the top 10 passages and we split them into two five-passage chunks.
We consider the $i$th passage as a positive passage if the predictions based on the passage collections including $i$th passage unanimously agree on correct prediction whereas it fails to generate the correct class when $i$th passage is masked. 
We consider the $i$th passage as a negative passage when (i) the model succeeds to answer when and only when $i$th passage is masked, and (ii) the predictions unanimously agree on incorrect classes, which indicates all of the passages do not support the input claim. 

\paragraph{Knowledge-enhanced dialogue.}
Unlike open-domain QA or fact verification, the final output of a dialogue system can be highly open-ended. 
For dialogue, we compare the average F1 score of the generated responses when $i$th passage is included and masked. 
If the average F1 when $p_i$ is presented is higher by more than 0.1 than the F1 when $p_i$ is masked, we consider $p_i$ provides useful evidence to generate the correct response, and therefore mark $p_i$ positive.
On the contrary, when the average F1 when $p_i$ is presented is lower by more than 0.1 than the score when $p_i$ is masked, we believe $p_i$ can be highly distracting, and thus we mark $p_i$ negative. 
As in fact verification, we use the top 10 passages and split them into two five-passage chunks.

\subsection{Implementation Details of Evidentiality Labeling Model}
We use PyTorch~\cite{paszke2019pytorch} via HuggingFace transformers RoBERTA~\cite{liu2019roberta} implementation.\footnote{\url{github.com/huggingface/transformers}} 
We tune our model from RoBERTa-base. 
We optimize the objective function using Adam~\cite{kingma2014adam} with learning rate $2 \times 10^{-5}$ .
We lowercase the input and set the maximum sequence length to 350. 
We train the model for 7 epochs. 
Per GPU batch size is 12 and we use 8 GPUs with 24 GB memory. 

\paragraph{Training data. }
We mine new training data for each task using our leave-one-out generation approach and mix the data with Natural Questions~\cite{kwiatkowski2019natural}.
For Natural Questions data, as human annotators annotate \texttt{long-answer}, from which final minimal answers are extracted, we assume that those human-annotated long answers are evidentiality-positive passages, while the other passages included in the same article and are not included in the long answers negative.
We first collect all of the \texttt{long-answer} passages from Natural Questions training data, and randomly sample two negative passages per questions with \texttt{long-answer} annotations. We discard the examples where long answers are list or table elements. 
Consequently, we obtain 250k training samples, and we use 90\% of the data as our training data and the remaining 10\% of the data as our development set.

\subsection{Examples of the Passages Mined by Leave-one-out Generation}
{\Tref{tab:loo_paragraphs_results} present several positive passages mined by leave-one-out generation approach. 
The positive passages for the open-domain QA and fact verification tasks clearly present the evidence leading to the gold answers (the highlighted sentences). 
Also in the first example of the knowledge-enhanced dialogue, the model finds a positive passage, which has high lexical overlap with the gold response. 
On the other hand, the second example shows the difficulty of finding the correct evidence for generation especially when the context history is long. The original dialogue history mentions skateboard and the last human utterance asks about when they were invented, while the passage labeled as positive is about electric skateboards and when they were released for the first time. 
We found due to the open-ended nature of knowledge-enhanced dialogue and F1 score-based positive passage labeling can be results in more false positive passages than other two tasks, as even the passage does not really support the evidence, it still helps a model generate a loosely grounded and related response and obtains higher F1 score. 
Recent work reports similar issues in long-form QA evaluations~\cite{krishna-etal-2021-hurdles}. 
}

\subsection{Examples of ${\mathbf{E}^{silver}}$ Obtained by $~\mathcal{M}$}
The newly mined examples can be seen in \Fref{img:new_evidentiality_example}. 
Although all of the passages here include gold answer strings, we observe that the red passages do not entail the answers. 
For instance, in the second example, the red passage from ``The Chronicles of Narnia: Prince Caspian'' only lists the names of the actors who reprise their roles from the first film, and does not mention show played ice queen. The first passage, on the other hand, clearly mentions that Tilda Swinton plays the White Witch (the ice queen) in the Chronicles of Narnia. 
The third example shows that our model detects the case where we originally have distantly-positive passages, all of which are labeled as negative by our evidentiality labeling model. 
The fourth example shows that the positive passages can be retrieved from multiple different articles, which are often not covered by existing datasets with gold paragraph annotations.

\section{Details of the Datasets}
\paragraph{License.}
Natural Questions~\cite{kwiatkowski2019natural}, TriviaQA~\cite{JoshiTriviaQA2017} is under Apache License 2.0. The KILT benchmark~\cite{petroni-etal-2021-kilt}, where our FEVER and Wizard of Wikipedia data is taken, is under MIT License. 
FAVIQ~\cite{park2021faviq} does not explicitly mention the license. 
We use all of the datasets for their intended uses.

\paragraph{Privacy-related information and harmful context.}
All of the datasets use the English Wikipedia or web articles as a knowledge source and the input queries are authored by human annotators, and we believe those resources are less likely to include personal information or harmful context. 
In addition, dataset creators often conduct intensive analysis on annotated data and discard problematic examples, which may further reduce the risk of the problematic content.

\section{More Analysis and Examples}
\subsection{Details of Task-specific heuristics for an ablation of $\mathbf{E}^{silver}$}
{
For open-domain QA, this model uses {answer string matching to supervise our multi-task learning.}
As discussed, this distantly supervised approach cannot be directly applied to classification or open-ended generation tasks. 
For WoW, it uses {\it provenance} title, which is the title of the Wikipedia article including the gold paragraph, and label all passages from provenance articles positive~\cite{petroni-etal-2021-kilt}. 
For FaVIQ-A, it uses the original answer annotations inherited from AmbigQA available in the dataset. 
It should be noted that that additional metadata is often unavailable in most of the datasets{, and this variant for WoW and FaVIQ can be considered as a ground-truth setting.}
}

\subsection{Analyzing Attentions of $\mathcal{G}$ and $\mathcal{G}^+$} 
\begin{figure}[h!]
  \includegraphics[width=0.9\linewidth]{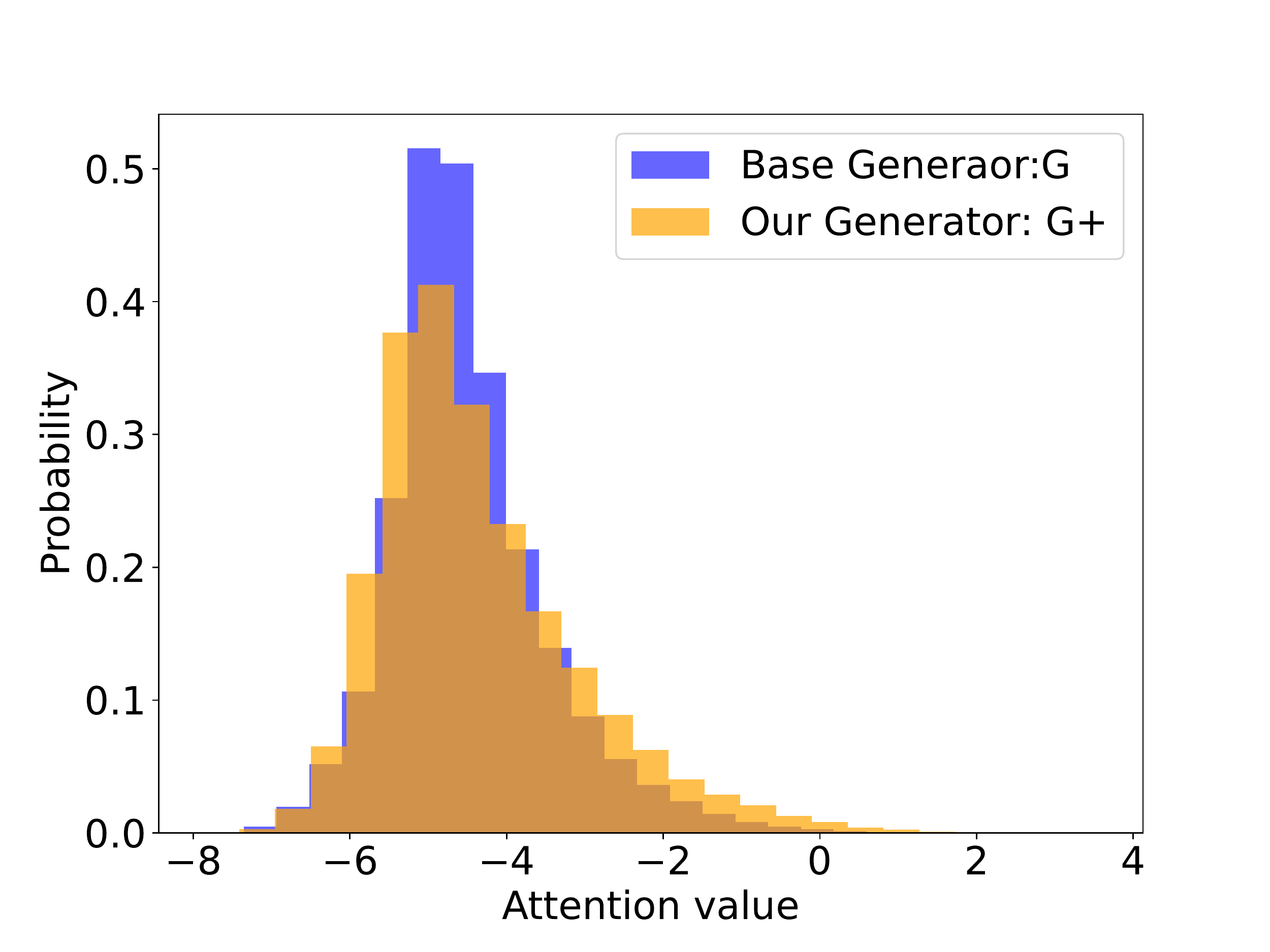}
  \caption{Attention score distributions over top 20 passages of the base generator $\mathcal{G}$ and our evidentiality-guided generator $\mathcal{G}^+$. 
  }
  \label{img:attention_analysis}
\end{figure}
To further understand our method's behavior, we compare the attention scores assigned to the top retrieved passages of a base generator FiD ($\mathcal{G}$) and our evidentiality-guided generator ($\mathcal{G}^+$). 
{\Fref{img:attention_analysis} shows that the attention scores of the base generator $\mathcal{G}$ and $\mathcal{G}^+$; the x-axis is the attention values and the y-axis is probability of the histogram. }
The attention scores of the base generator $\mathcal{G}$ are concentrated closely near the value of -5.0, whereas the attention scores of our $\mathcal{G}^+$  more widely spread out. 
We also found that our $\mathcal{G}^+$ more often gives its highest attention value to the passages ranked lower by $\mathcal{R}$; our generator $\mathcal{G}^+$ and base generator $\mathcal{G}$ gives their highest attention scores to the passages ranked lower than top 10 by $\mathcal{R}$ in 45.8\% and 44.8\% of the examples, respectively.
{
We hypothesize that FiD mostly generates answers from more highly-ranked passages while our method enables shifting the attention scores to lower-ranked passages and generates answers from those, by explicitly training the models telling the evidentiality-negative and evidentiality-positive passages.  
}

\subsection{Examples from Qualitative Analysis on FaVIQ-A}
\Tref{tab:qual_faviq} shows the most attended passages and final prediction results made by the base generator $\mathcal{G}$ (FiD) and our evidentiality generator $\mathcal{G}^+$ (ours) from our qualitative analysis on FaVIQ-Ambig.

\begin{table*}[ht!]
\center
\begin{tabular}{p{0.95\linewidth}}
\toprule 
{\it {\bf Category 1 (40\%)}: Our model attends a more relevant passage.} \\\midrule
{\bf Claim}: roger danuarta was the name of actress in munna michael as judge of dancing stars from jodhpur, rajasthan, india.\\
{\bf A}: REFUTES
\\\hline
\multirow{3}{\linewidth}{ {\bf [Ours (pred: REFUTES)] Munna Michael}:~~as Judge of Dancing Star (cameo appearance) Chitrangada Singh as Judge of ``Dancing Star'' (cameo appearance) Pallavi Kulkarni (cameo appearance) \hl{Roger Danuarta} (cameo appearance)}\\\\
\\\hdashline
\multirow{4}{\linewidth}{ {\bf [FiD (pred: SUPPORTS)] Dancing with the Stars (American season 24)}:~~Dancing with the Stars (American season 24) The full list of celebrities and pros was announced on March 1, 2017, on Good Morning America. Hosts and judges. Tom Bergeron and Erin Andrews returned as hosts, and Carrie Ann Inaba, Len Goodman, Julianne Hough, and Bruno Tonioli returned as judges}\\
\\
\\
\\
\toprule 
{\it {\bf Category 2 (10\%)}: FiD attends a more relevant passage.} \\\midrule
{\bf Claim}: west was stacey's surname in gavin and stacey before marrying. \\
{\bf A}: SUPPORTS
\\\hline
\multirow{4}{\linewidth}{ {\bf [Ours (pred: REFUTES)] List of Gavin \& Stacey characters}:~~``Gavin \& Stacey'' is an award winning British television comedy series, following the lives of the title characters Gavin (Mathew Horne) and Stacey (Joanna Page), who, before marrying, live on opposite sides of the country, Gavin in Billericay, Essex, and Stacey in Barry, Vale of Glamorgan. }\\
\\
\\
\\\hdashline
\multirow{4}{\linewidth}{ {\bf [FiD (pred: SUPPORTS)]Gavin \& Stacey}:~~Gavin \& Stacey Other storylines that run throughout the course of the three series include Pam\'s fake vegetarianism. Characters and cast.:Main characters. Gavin Shipman (Mathew Horne) – nicknamed ``Gav'', ``Gavlar'', or ``Gavalar'', the funny and enthusiastic level-headed protagonist from Billericay, Essex. Stacey Shipman (``nee'' \hl{West})} \\
\\
\\
\\
\toprule 
{\it {\bf Category 3 (30\%)}: Both are equally irrelevant.} \\\midrule
{\bf Claim}: sylvia fricker was the original singer of always on my mind. \\
{\bf A}: SUPPORTS
\\\hline
\multirow{3}{\linewidth}{ {\bf [Ours (pred: SUPPORTS)] For Once in My Life (Sylvia Syms album)}:~~`For Once in My Life (Sylvia Syms album) For Once in My Life is an album by American vocalist Sylvia Syms recorded in 1967 and released on the Prestige label. }\\
\\
\\\hdashline
\multirow{3}{\linewidth}{ {\bf [FiD (pred: REFUTES)]Follow Me...}:~~Follow Me... The song ``You Were on My Mind'' was originally recorded and released in 1964 by Ian \& Sylvia, and was a major hit in the US when covered by the group We Five in 1965. } \\
\\\\
\toprule 
{\it {\bf Category 4 (20\%)}: Both are equally relevant.} \\\midrule
{\bf Claim}: the third party system ended in american politics in 1854. \\
{\bf A}: REFUTES
\\\hline
\multirow{3}{\linewidth}{ {\bf [Ours (pred: REFUTES)]Political parties in the United States}:~~The GOP dominated national politics during the Third Party System, \hl{from 1854 to 1896}, and the Fourth Party System from 1896 to 1932. }\\
\\
\\\hdashline
\multirow{3}{\linewidth}{ {\bf [FiD (pred: SUPPORTS)] Third Party Syste}:~~The Third Party System is a term of periodization used by historians and political scientists to describe the history of political parties in the United States \hl{from 1854 until the mid-1890s}. } \\
\\
\\\bottomrule
\end{tabular}
\caption{Examples of the most attended passages and final prediction results made by the base generator $\mathcal{G}$ (FiD) and our evidentiality generator $\mathcal{G}^+$ (ours) from our qualitative analysis on FaVIQ-Ambig.}
\label{tab:qual_faviq}
\end{table*}

\end{document}